\documentclass{llncs}
\usepackage{multirow}
\usepackage{cite}
\usepackage{float}
\usepackage{graphicx}
\usepackage{xcolor}
\graphicspath{{images/}}
\usepackage{caption}
\usepackage{subcaption}
\begin{document}

\title{Bridging between Computer and Robot Vision through Data Augmentation: a Case Study on Object Recognition}
\titlerunning{}  
%
\author{Antonio D'Innocente, Fabio Maria Carlucci, Mirco Colosi, Barbara Caputo}
\maketitle

\begin{abstract}
Despite the impressive progress brought by deep network in visual object recognition, robot vision is still far from being a solved problem. The most successful convolutional architectures are developed starting from ImageNet, a large scale collection of images of object categories downloaded from the Web. This kind of images is very different from the situated and embodied visual experience of robots deployed in unconstrained settings. To reduce the gap between these two visual experiences, this paper proposes a simple yet effective data augmentation layer that zooms on the object of interest and simulates the object detection outcome of a robot vision system. The layer, that can be used with any convolutional deep architecture, brings to an increase in object recognition performance of up to 7\%, in experiments performed over three different benchmark databases. Upon acceptance of the paper, our robot data augmentation layer will be made publicly available.
\end{abstract}
\section{Introduction}

 The ability to understand what they see is crucial for autonomous robots deployed in unconstrained settings, such as those shared with humans. Recent advances in visual recognition, induced by the deep learning tidal wave, has brought high hopes that the very same impressive progresses seen in the computer vision community would have been quickly shared by the robot vision community\cite{dlMoboro}. Experimental evaluations have repeatedly shown that this is not the case. Although the use of convolutional neural networks  has brought important improvements in performance, compared to approaches based on shallow classifiers, several authors have shown that we are still far from the level of performance necessary to robots in the wild (we refer to section \ref{related} for a review of previous work on the topic). 

\begin{figure}[htb]
\centering
\begin{subfigure}{.5\textwidth}
  \centering
  \includegraphics[width=.8\linewidth]{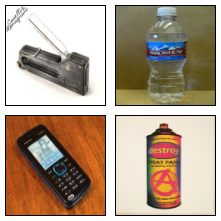}
\end{subfigure}%
\begin{subfigure}{.5\textwidth}
  \centering
  \includegraphics[width=.8\linewidth]{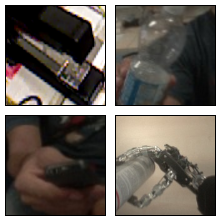}
\end{subfigure}
\caption{Images from classes \textit{stapler, water bottle, cellphone, spray can} as seen in \textit{Imagenet} (left) and \textit{JHUIT-50, HelloiCubWorld} (right)}
\label{fig:comparison}
\end{figure}

An issue that has called considerable attention is the difference between web images and robot images (figure \ref{fig:comparison}). Web images, that constitute the main training resource of modern deep networks, tend to show objects in the center of the scene, in various contexts, from canonical view-points, i.e. view-points capturing the most informative parts of the object of interest (figure \ref{fig:comparison}, left). As opposed to this, robots acquire snapshots of objects to be recognised based on the figure-ground segmentation algorithms they are equipped with. This often leads to objects imaged at unusual angles and scale, and only partially visible in the image (figure \ref{fig:comparison}, right). This is a crucial issue, because the overwhelming majority of deep visual recognition networks are trained over ImageNet, a 1.4M images database of 1000 object categories derived from the Web \cite{imagenet:ilsvrc}. Hence, any visual object recognition system attempting to use such deep networks for robot vision, is attempting to recognise objects based on a very different type of visual information than what the robot perceives in its situated scenario.

The focus of this paper is on how to bridge among these two different visual domains, with the aim to increase the performance of deep visual object recognition networks when used on robotic data. We propose to enrich the original web images with rescaled and cropped version of the original view, so to simulate to a certain extent the visual experience of an autonomous agent.  The procedure can be integrated into any deep architecture as a data augmentation layer. Extensive experiments on $3$ different databases and $2$ different deep networks show that our approach  leads to increases in absolute performance of up to 7\%. Upon acceptance of the paper, we will release a \texttt{python} implementation for a seamless integration of our data augmentation strategy in any deep network.

\begin{figure}[htb]
  \centering
    \includegraphics[width=1.0\textwidth]{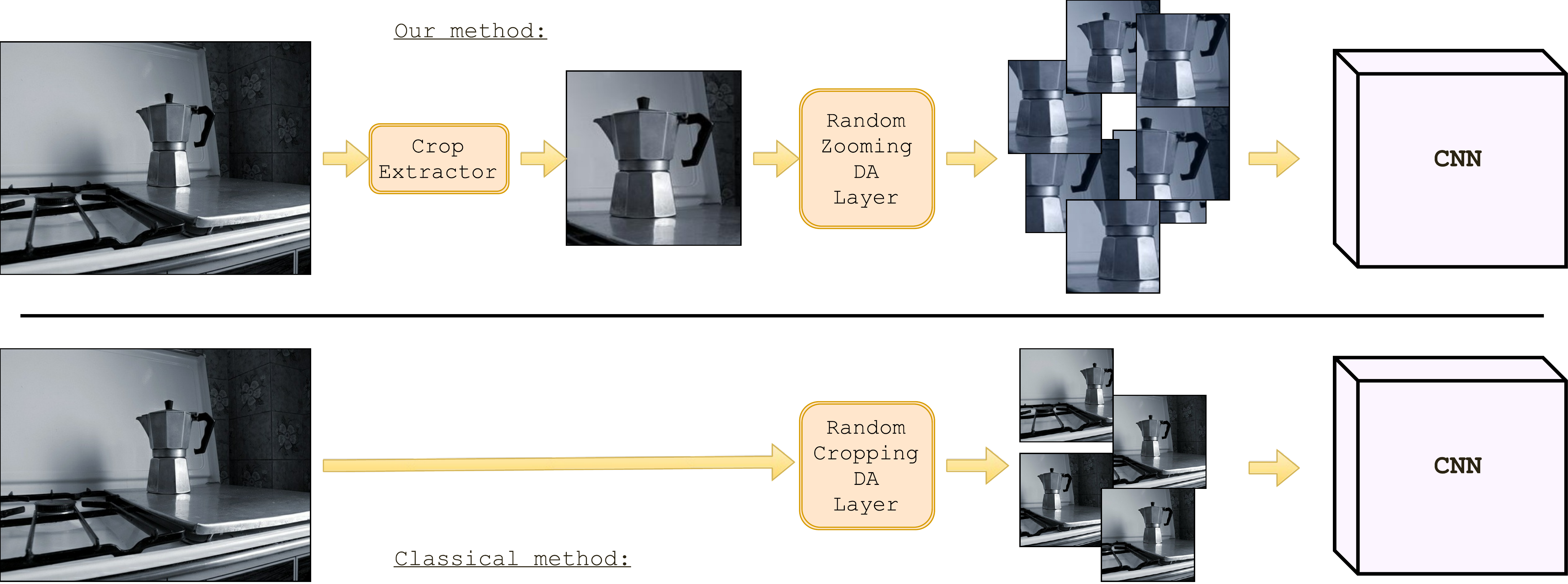}
   \caption{Above, the proposed pipeline: from each image we extract one or more objects, using bounding box annotations. Each centered object goes then through our data augmentation layer which extracts a number of crops at varying zoom levels. Below, the classical\cite{alexnet} approach: each image is randomly cropped by a small amount, with no notion of objectness.}
   \label{fig:architecture}
\end{figure}

The rest of the paper is organized as follows: after a review of relevant literature (section \ref{related}), we describe our strategy for generating robot-like object images starting from web images, and how the data augmentation layer works in practice. Section \ref{experiments} reports our experimental setup and the results obtained over the Washington-RGBD\cite{washington}, JHUIT-50\cite{jhuit} and HelloiCubWorld \cite{helloicub} databases. We conclude with a summary and discussing future research directions. 
\section{Related Work}
\label{related}
State of the art work in computer vision reveals that deep convolutional models, when pre-trained on a large and diverse dataset like ImageNet \cite{imagenet}, are able to extract general and high level information from images \cite{decaf}. Exploiting this generalized knowledge for new tasks is common practice. In \cite{semantic}, the use of a pre-trained convolutional model, installed on a robot, was combined with context aware semantic web mining for object recognition. In \cite{pose}, a pre-trained CNN model was used as feature extractor on the RGB-D Object dataset \cite{washington} for pose estimation. Large convolutional models pre-trained on RGB images have been used to extract rich features from depth images by representing depth with three channels \cite{rich}.  \cite{multimodal} combined RGB and depth classification, parallelizing deep pre-trained models for each modality. Robot images are usually more prone to scaling and translation noise; \cite{pasquale} shows behaviours of features extracted from pre-trained deep models with different degrees of visual transformations. While the robot vision community is exploring several valid strategies, recent works in the field  share the same specific step: they use an AlexNet \cite{alexnet} model pre-trained on the ImageNet 1000 object categories dataset \cite{imagenet:ilsvrc} for classification or feature extraction on RGB images. We argue that features extracted from these models, while having shown generality \cite{decaf}, are still tied to the original representations, suffering from their own bias. Web images are directly downloaded from the web, and often they have been acquired by humans and subjected to manual cleaning and annotation. Hence, they are prone to heavy background noise. Robot acquired images are often  taken in large quantities in the same setting instead, such as the same room or office, and are subject to a different kind of bias. The robot often walks around the office, taking pictures of various things of interest. In our approach we train deep convolutional models by artificially injecting robot images' bias on the training dataset. By applying bounding box cropping and random zooming transformations on web images, we make them more similar to robot images and subsequently test how our adapted models perform in robot vision task against models trained on original images. We are not aware of any previous work in this direction.
\section{Robot Specific Data Augmentation}

We propose a methodology for improving performances in robot vision tasks when pre-training deep convolutional models on web images.

Very often CNN models pre-trained on the 1000 object categories dataset \cite{imagenet:ilsvrc} are used for feature extraction on robotic tasks. Classification scores in these scenarios are good but not exceptional. The state of the art seems to suggest problems related to a domain gap between ImageNet's \cite{imagenet} web acquired images and robot images. Web images are often acquired by humans and the main source of noise is the background, which may vary significantly between pictures belonging to the same category. Robot acquired images are subject to a different kind of noise, objects are usually zoomed in or translated, while the background view is limited.

Since pre-trained models are trained on web pictures, we process the CNN's training dataset to make its images resemble robot-acquired images, so that our deep convolutional models learn features more resistant to robot vision noise. Instead of training directly on web images, we use off-line preprocessing to build a crop dataset by using bounding boxes annotation. We also enlarge original bounding boxes by 20\% before cropping. At training time, we use random zooming on the object crops to make the network see randomly zoomed versions of the same object. The data-augmentation extracts patches at casual position, with a random zooming factor between 1.0x and 2.0x. After zooming, images got resized to fit the first convolutional layer's input size using bilinear interpolation.

Combinining our off-line and on-line processing methodologies we trained networks on patches containing zoomed-in or partially excluded objects, common properties for robot images, and results show that models pre-trained on these object parts significantly outperform models pre-trained on original images in every robot vision task we tested them for.
\section{Experiments}
\label{experiments}
We train two models for each of our training datasets, a model based on the "AlexNet" architecture \cite{alexnet} and a model based on the "Inception-v3" architecture \cite{inception-v3}, using the data-augmentation techniques described in previous section. We then select several robotic datasets with centered, non-centered, zoomed-in and artificially translated images, and use our pre-trained CNNs models as features extractors on these datasets \cite{decaf}. Lastly, we run a linear classifier on the extracted features to evaluate how well our models have generalized to the robotic tasks.

\subsection{Databases}

\subsubsection{CNN's training datasets.} We collect 3 datasets from ImageNet \cite{imagenet} for training our models. We will refer to these datasets as Baseline, Clean Crops and Dirty Crops for the rest of the article. The Baseline dataset consists of a set of images, all having bounding boxes annotations, for which, like in the 1000 object categories dataset \cite{imagenet:ilsvrc}, there is no semantic overlap between the classes. Clean Crops is another dataset without semantic overlaps between the classes, it contains 930.000 images, obtained from Baseline by cropping objects outside original pictures using bounding boxes annotation. For the Dirty Crops dataset, we used every ImageNet's bounding box to crop objects, and as a result, this dataset contains many IS-A relationships between its classes, but, with approximately 1.2 million images, its also larger than the Clean Crops.

\subsubsection{Robotic datasets.} 
\begin{figure}[htb]
\centering
\includegraphics[scale=0.5]{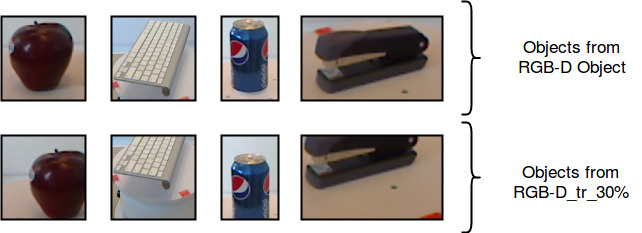}
\caption{Original crops from the RGB-D Object dataset (top) and the same crops after random translations (bottom)}
\label{fig:dirty_washington}
\end{figure}

We run our experiments on the JHUIT-50 \cite{jhuit}, HelloiCubWorld \cite{helloicub} and RGB-D Object \cite{washington} datasets. The JHUIT-50 dataset is focused on the task of fine-grained instance recognition and consists of 50 objects and hand tools, with images being acquired by rotating a camera around the objects at different heights - it contains almost $15.000$ images. HelloiCubWorld contains a "human" and a "robot" dataset. Images in the datasets are obtained by using human or robot modes of acquisition. In the human mode acquisition, a human moves the object while the robot tracks it. In the robot mode acquisition, the robot moves the object in its own hand, tracking it. The HelloiCubWorld-human dataset is  obtained by using human mode acquisition, and the HelloiCubWorld-robot dataset is obtained by using robot mode acquisition. Both datasets consist of objects organized in 7 classes and each of them contain $7.000$ images. The RGB-D Object dataset has been obtained by collecting frames from objects spinning on a turntable, it consists of 300 objects organized into 51 categories; with around $50.000$ images it's the largest we experiment on. We also created artificially translated versions of the RGB-D Object dataset to study translation invariance of features extracted by our models. We created 3 additional versions, RGB-D\_tr\_10\%, RGB-D\_tr\_20\% and RGB-D\_tr\_30\% for which we have randomly translated original crops by 10\%, 20\% and 30\% respectively (see fig. \ref{fig:dirty_washington}).

\subsection{Architectures}

\subsubsection{AlexNet.} AlexNet \cite{alexnet} is the architecture behind the competition winning model for the ILSVRC 2012 competition \cite{imagenet:ilsvrc}, it consists of 5 convolutional layer, followed by 3 fully connected layer. The AlexNet is currently one of the most widely used architecture in the robotic community.

\subsubsection{Inception-v3.} The Inception-v3 \cite{inception-v3} is an improved version of the original Inception architecture \cite{inception}. It consists of several inception modules stacked on top of each other. Compared to the AlexNet \cite{alexnet}, the Inception-v3 is a more advanced architecture which achieved far better classification results on the ILSVRC competition.

\subsection{Results}

For each architecture, we trained one model on the Baseline dataset with random cropping, and two models for each of the crops datasets, one with random cropping and the other with random zooming. On AlexNet, random cropping was obtained by taking random 227x227 patches from the 256x256 input images; on Inception-v3, we took 299x299 patches on 384x384 input images. AlexNet models were trained by setting initial learning rate at 0.01 and dividing it by a factor of 3 each time validation loss didn't improve for 4 consecutive epochs; this procedure was repeated 6 times. Inception-v3 models were trained by setting learning rate at 0.01 and dividing it by a factor of 10 each time validation loss got worse, with this procedure being repeated 2 times. All models were also trained using random horizontal flipping.

After training, we extracted features of robotic images from different layers. On the AlexNet models, we extracted features from pool5, fc6 and fc7. On the Inception-v3 models, we extracted features from a max pooling layer and an average pooling layer we placed on top of the last convolution; for both of those layers we used kernel size 8 and stride 2.

Final classification occurred by running a linear SVM on extracted features. Hyperparameter for the SVM was chosen by running a 3-fold cross validation for the RGB-D Object dataset and its variations, while it was left at the default value for the other robotic datasets.

Model's results are summarized in the following tables and charts. Accuracy score refers to results obtained by running the linear classifiers on features extracted by a model trained on a specific dataset and with a corresponding data-augmentation technique. Each Table refers to features extracted from a certain architecture, which is indicated in the caption.

\begin{table}[htb]
\centering
\caption{SVM results using AlexNet features on the HelloiCubWorld and JHUIT-50 datasets. In bold the best results for a given layer. Best overall results are highlighted in red.}
\label{table:alexnet-all}
\begin{tabular}{cc|ccc|c}
\hline
\begin{tabular}[x]{@{}c@{}}Training\\dataset\end{tabular} & \begin{tabular}[x]{@{}c@{}}Data\\augmentation\end{tabular} & \begin{tabular}[x]{@{}c@{}}iCub\\(human)\end{tabular} & \begin{tabular}[x]{@{}c@{}}iCub\\(robot)\end{tabular} & JHUIT-50 & Layer \\ \hline \hline
Baseline                                    & Random Cropping                                          & $93.66$                           & $94.11$                           & $85.00$           & \multirow{5}{*}{pool5}            \\
Clean Crops                                & Random Cropping                                          & $94.14$                           & $\mathbf{96.31}$                           & $85.36$                 &      \\
Clean Crops                                 & Random Zooming                                           & $93.49$                           & $96.09$                           & $85.79$               &        \\
Dirty Crops                                 & Random Cropping                                          & $93.66$                           & $94.11$                           & $85.00$                 &      \\
Dirty Crops                                 & Random Zooming                                           & $\mathbf{94.60}$                           & $95.60$                           & $\color{red} \mathbf{87.05}$   & \\
\hline
Baseline                                  & Random Cropping                                          & $91.91$                           & $92.91$                           & $84.53$            &           \multirow{5}{*}{fc6}\\
Clean Crops                                 & Random Cropping                                         & $93.09$                           & $94.00$                           & $84.13$                    &   \\
Clean Crops                                 & Random Zooming                                           & $94.26$                           & $\mathbf{97.54}$                           & $84.04$                   &    \\
Dirty Crops                                & Random Cropping                                          & $93.27$                           & $94.91$                           & $84.67$                   &    \\
Dirty Crops                                 & Random Zooming                                           & $\color{red} \mathbf{94.86}$                           & $96.89$                           & $\mathbf{85.41}$ &  \\
\hline
Baseline                                    & Random Cropping                                          & $89.94$                           & $90.66$                           & $\mathbf{81.82}$        &        \multirow{5}{*}{fc7}\\
Clean Crops                                 & Random Cropping                                          & $90.46$                           & $93.43$                           & $81.58$                   &    \\
Clean Crops                                 & Random Zooming                                           & $92.86$                           & $\color{red} \mathbf{97.89}$                           & $81.18$                   &    \\
Dirty Crops                                  & Random Cropping                                          & $\mathbf{93.46}$                           & $95.17$                           & $81.62$                 &      \\
Dirty Crops                                  & Random Zooming                                           & $93.23$                           & $96.11$                           & $81.09$   &  \\
\hline
\end{tabular}
\end{table}

\begin{table}[htb]
\centering
\caption{SVM results using AlexNet features on the RGB-D Object dataset and it's artificially translated versions. In bold the best results for a given layer. Best overall results are highlighted in red.}
\label{table:alexnet-tr}
\resizebox{\textwidth}{!}{
\begin{tabular}{cc|cccc|c}
\hline
\begin{tabular}[x]{@{}c@{}}Training\\dataset\end{tabular} & \begin{tabular}[x]{@{}c@{}}Data\\augmentation\end{tabular} & RGB-D Object & tr\_10\% & tr\_20\% & tr\_30\% & Layer \\ \hline \hline
Baseline                                    & Random Cropping                        & $84.90 \pm 1.02$                  & $82.92 \pm 1.17$                & $79.41 \pm 1.30$                & $76.74 \pm 1.53$            & \multirow{5}{*}{pool5}            \\
Clean Crops                                 & Random Cropping                        & $86.29 \pm 1.83$                  & $84.85 \pm 1.56$                & $81.85 \pm 1.81$                & $79.68 \pm 1.75$            &    \\
Clean Crops                                & Random Zooming                         & $88.38 \pm 1.42$                  & $86.55 \pm 1.63$                & $83.64 \pm 1.56$                & $81.70 \pm 1.62$            &    \\
Dirty Crops                                 & Random Cropping                        & $87.53 \pm 1.97$                  & $85.79 \pm 1.97$                & $82.88 \pm 1.89$                & $80.49 \pm 1.76$           &     \\
Dirty Crops                                  & Random Zooming                         & $\color{red} \mathbf{88.67 \pm 1.81}$                  &  $\color{red} \mathbf{87.13 \pm 2.07}$                & $\color{red} \mathbf{84.71 \pm 2.07}$                & $\color{red} \mathbf{82.31 \pm 2.07}$   & \\
\hline
Baseline                                    & Random Cropping                        & $81.99 \pm 1.20$                  & $80.28 \pm 1.36$                & $77.66 \pm 1.32$                & $75.39 \pm 1.43$            &           \multirow{5}{*}{fc6}\\
Clean Crops                              & Random Cropping                        & $84.38 \pm 2.00$                  & $82.94 \pm 2.01$                & $80.36 \pm 1.76$                & $77.74 \pm 1.70$             &   \\
Clean Crops                                & Random Zooming                         & $85.36 \pm 1.59$                  & $83.21 \pm 2.43$                & $81.82 \pm 2.03$                & $80.41 \pm 1.89$            &    \\
Dirty Crops                                  & Random Cropping                        & $85.76 \pm 2.22$                  & $84.11 \pm 2.31$                & $80.90 \pm 2.10$                & $79.02 \pm 1.81$            &    \\
Dirty Crops                                  & Random Zooming                         & $\mathbf{86.05 \pm 1.99}$                  & $\mathbf{84.54 \pm 2.29}$                & $\mathbf{82.75 \pm 2.02}$                & $\mathbf{80.93 \pm 1.74}$ &  \\
\hline
Baseline                                    & Random Cropping                        & $78.15 \pm 1.40$                  & $76.62 \pm 1.36$                & $74.08 \pm 1.31$                & $71.13 \pm 71.13$        &        \multirow{5}{*}{fc7}\\
Clean Crops                                 & Random Cropping                        & $81.63 \pm 2.39$                  & $80.07 \pm 2.55$                & $76.84 \pm 2.29$                & $73.68 \pm 2.23$           &     \\
Clean Crops                                 & Random Zooming                         & $81.80 \pm 1.93$                  & $80.64 \pm 2.16$                & $78.90 \pm 2.08$                & $77.04 \pm 2.12$          &      \\
Dirty Crops                                  & Random Cropping                        & $\mathbf{82.93 \pm 2.32}$                  & $80.80 \pm 2.25$                & $77.53 \pm 2.05$                & $74.73 \pm 1.97$          &      \\
Dirty Crops                                  & Random Zooming                         & $82.58 \pm 2.54$                  & $\mathbf{81.57 \pm 2.47}$                & $\mathbf{79.97 \pm 2.56}$                & $\mathbf{77.63 \pm 2.00}$    &  \\
\hline
\end{tabular}}
\end{table}

\begin{figure}[htb]
  \centering
    \includegraphics[width=1.0\textwidth]{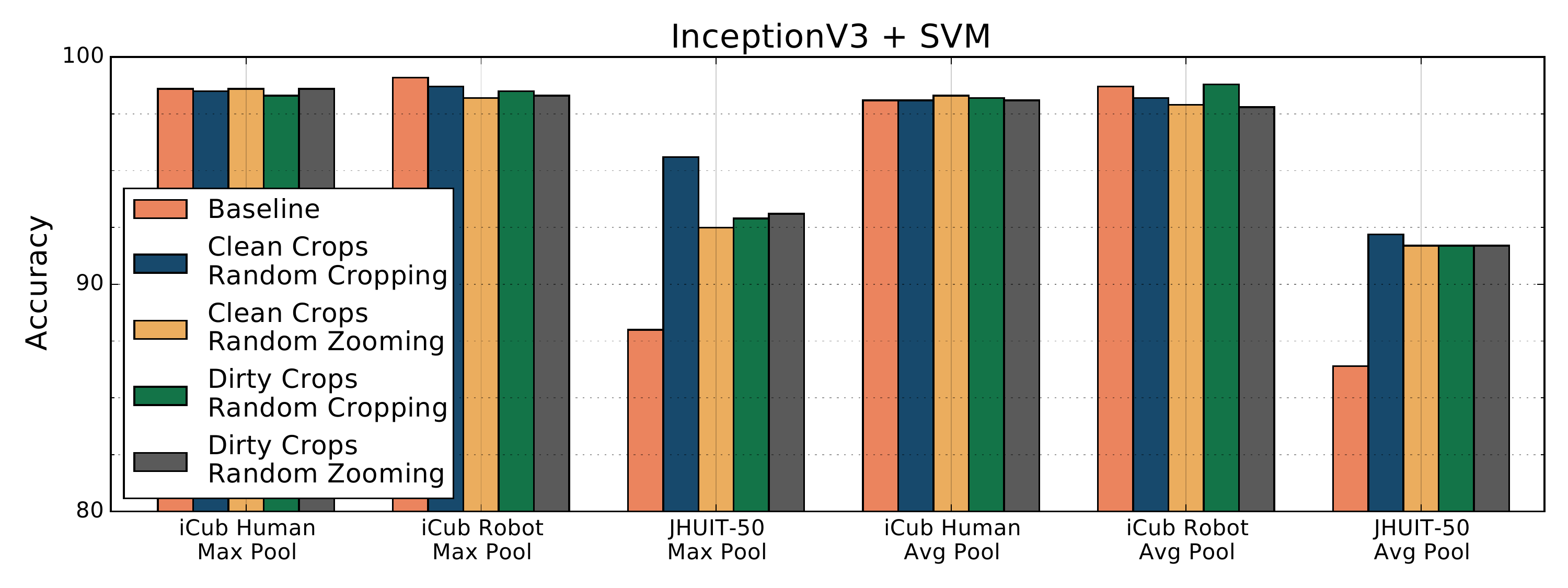}
   \caption{Results on HelloiCubWorld and JHUIT-50 - InceptionV3. Notice how our proposed methods boosts extensively the accuracy on JHUIT-50; the same cannot be said about HelloiCubWorld but it must be kept in mind that it is a small dataset, the accuracy is already extremely high, and the differences are statistically insignificant.}
   \label{fig:washingtonIncV3All}
\end{figure}

\begin{figure}[htb]
  \centering
    \includegraphics[width=1.0\textwidth]{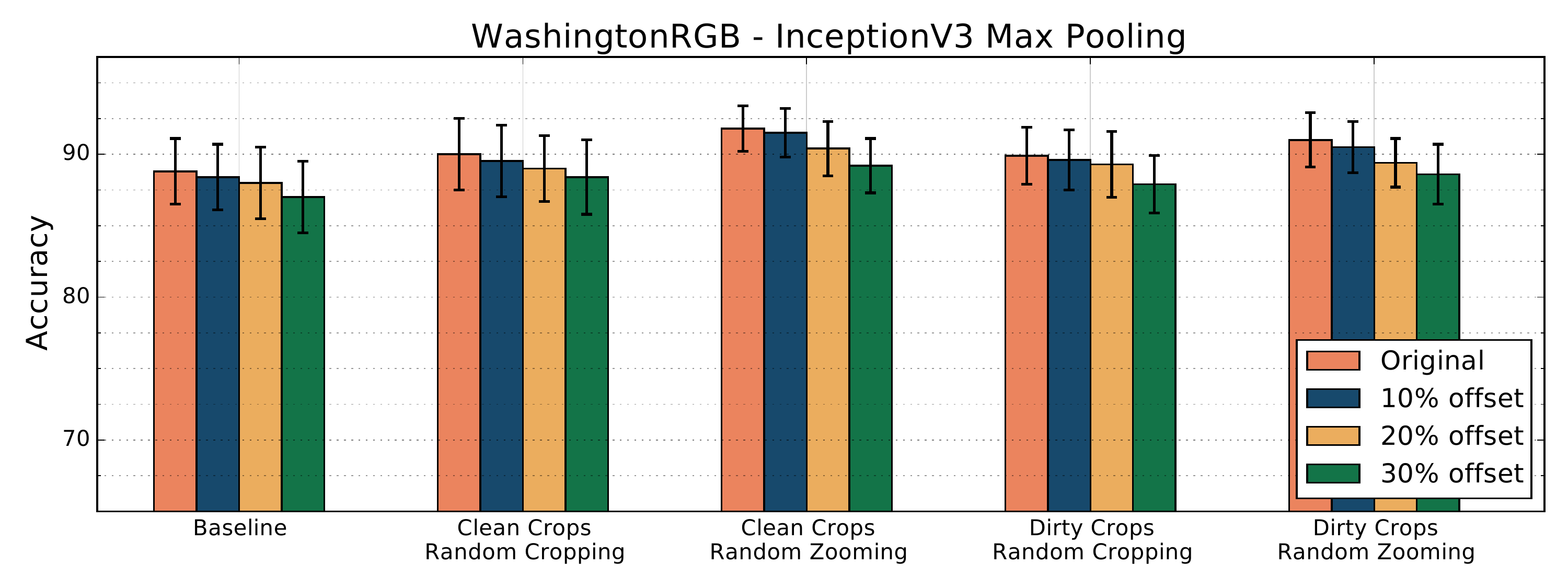}
   \caption{Results on Washington RGB - InceptionV3 Max Pool.}
   \label{fig:washingtonIncV3MaxPool}
\end{figure}

On Inception-v3 models, features extracted from the average pooling layer provided inferior results compared to features extracted from the max pooling layer for the JHUIT-50 and HelloiCubWorld datasets. For this reason we didn't test the average pooling's features on the RGB-D Object datasets and its variations.

\subsection{Discussion}

Our results show that the best models are those trained on the crops datasets using random zooming. On the AlexNet architecture the best model has been the one trained on Dirty Crops, while the Inception-v3 models trained on Dirty Crops and Clean Crops had comparable results, suggesting that AlexNet generalized better features by using more training images, even in the presence of semantically overlapping labels.

Inception-v3 model trained on Clean Crops achieved a mean accuracy score of $91.76$ on the RGB-D Object splits, compared to the Baseline model which scored $88.78$, and also scored $92.50$ points on the JHUIT-50 task, 4.5 points higher than the corresponding Baseline score. Inception-v3 model trained on Dirty Crops also surpassed the Baseline on the RGB-D Object and JHUIT-50 tasks, scoring $91.04$ points and $93.13$ points respectively. Models trained with our methodology and tested on the HelloiCubWorld-human dataset didn't improve the Baseline results, while they performed worse for the HelloiCubWorld-robot task, albeit the difference is lower than one point and the dataset is small.

We also test the models we trained on artificially translated versions of the RGB-D Object dataset. Features extracted from the models trained on Crop datasets with Random Zooming performed better than features extracted from models trained on the Baseline against increasing levels of artificial translations.

Since we employed Random Zooming to simulate the zooming and off-target detection phenomena we commonly see in robotic data-set, we used it only for models trained on object crops. Since we are comparing models trained with random zooming with a baseline model trained with random cropping, it could be argued  that it is the random zooming itself giving better results, and not the crop dataset used for training. Also the Dirty Crops datasets contain crops extracted from images not in the Baseline dataset. However, by comparing results obtained with the Baseline models with results obtained with the models trained on Clean Crops with random cropping, we observe that using the crop dataset is also beneficial for the final classification task. Features extracted from the AlexNet models trained on Clean Crops with random cropping outperformed the corresponding Baseline features in every task. Features extracted from the Inception-v3 model trained on Clean Crops with random cropping outperformed Baseline features by $1.23$ and $4.57$ points on the RGB-D Object and JHUIT-50 tasks, while they got outperformed on the smaller HelloiCubWorld human and robot datasets by $0.08$ and $0.5$ points.
\section{Conclusions}
This paper presented a simple yet effective approach for improving the object classification accuracy of any deep network trained over large scale databases collected from the Web and used in the robot vision context. Our idea is to increase the similarity between the data acquired in the two visual domains by randomly scaling, zooming and cropping each image in a data augmentation layer. Results over three different benchmark databases confirm the effectiveness of the method.

While this technique certainly give a small but consistent help in bridging between computer and robot vision, very many challenges remain open. In particular, we will dedicate future work attempting to close the gap between these two different perceptual tasks by leveraging over the domain adaptation literature, aiming for methods working in the unsupervised domain adaptation setting without heavy requirements about the deep architecture of choice.
%
%

\end{document}